\documentclass{article}
\usepackage{arxiv}

\usepackage[utf8]{inputenc} 
\usepackage[T1]{fontenc}    
\usepackage{hyperref}       
\usepackage{url}            
\usepackage{booktabs}       
\usepackage{amsfonts}       
\usepackage{nicefrac}       
\usepackage{microtype}      
\usepackage{graphicx}
\usepackage[numbers]{natbib}
\usepackage{doi}

\usepackage[auth-lg]{authblk}

\title{Structured Prompt Interrogation and Recursive Extraction of Semantics (SPIRES): a method for populating Knowledge Bases using zero-shot learning}

\date{December 22, 2023}	

\author[1]{J. Harry Caufield}
\author[1]{Harshad Hegde}
\author[2]{Vincent Emonet}
\author[1]{Nomi L. Harris}
\author[1]{Marcin Joachimiak}
\author[3]{Nicolas Matentzoglu}
\author[4]{HyeongSik Kim}
\author[1]{Sierra Moxon}
\author[1]{Justin T. Reese}
\author[5]{Melissa A. Haendel}
\author[6]{Peter N. Robinson}
\author[1]{Christopher J. Mungall}

\affil[1]{Division of Environmental Genomics and Systems Biology,
Lawrence Berkeley National Laboratory, Berkeley, CA 94720, USA}
\affil[2]{Institute of Data Science, Faculty of Science and Engineering, Maastricht University, Maastricht, The Netherlands}
\affil[3]{Semanticly Ltd, Athens, Greece}
\affil[4]{Robert Bosch LLC, Sunnyvale, CA 94085, USA}
\affil[5]{Anschutz Medical Campus, University of Colorado, Aurora, CO 80217, USA}
\affil[6]{The Jackson Laboratory for Genomic Medicine, Farmington, CT 06032, USA}

\renewcommand{\headeright}{}
\renewcommand{\undertitle}{}
\renewcommand{\shorttitle}{SPIRES}

\hypersetup{
pdftitle={Structured Prompt Interrogation and Recursive Extraction of Semantics (SPIRES): a method populating Knowledge Bases using zero-shot learning},
pdfsubject={},
pdfauthor={},
pdfkeywords={large language models, linked data, schemas, text mining},
}

\begin{document}
\maketitle

\begin{abstract}
Creating knowledge bases and ontologies is a time consuming task that relies on manual curation. AI/NLP approaches can assist expert curators in populating these knowledge bases, but current approaches rely on extensive training data, and are not able to populate arbitrarily complex nested knowledge schemas. \\
Here we present Structured Prompt Interrogation and Recursive Extraction of Semantics (SPIRES), a Knowledge Extraction approach that relies on the ability of Large Language Models (LLMs) to perform zero-shot learning (ZSL) and general-purpose query answering from flexible prompts and return information conforming to a specified schema. Given a detailed, user-defined knowledge schema and an input text, SPIRES recursively performs prompt interrogation against an LLM to obtain a set of responses matching the provided schema. SPIRES uses existing ontologies and vocabularies to provide identifiers for matched elements. \\
We present examples of applying SPIRES in different domains, including extraction of food recipes, multi-species cellular signaling pathways, disease treatments, multi-step drug mechanisms, and chemical to disease orange relationships. Current SPIRES accuracy is comparable to the mid-range of existing Relation Extraction (RE) methods, but greatly surpasses an LLM's native capability of grounding entities with unique identifiers. SPIRES has the advantage of easy customization, flexibility, and, crucially, the ability to perform new tasks in the absence of any new training data. This method supports a general strategy of leveraging the language interpreting capabilities of LLMs to assemble knowledge bases, assisting manual knowledge curation and acquisition while supporting validation with publicly-available databases and ontologies external to the LLM. \\
SPIRES is available as part of the open source OntoGPT package: 
\url{https://github.com/monarch-initiative/ontogpt}. \\
\textbf{Contact:} \href{jhc@lbl.gov}{jhc@lbl.gov}\\

\end{abstract}

\section{Introduction}\label{sec1}
Knowledge Bases and ontologies (here collectively referred to as KBs) encode domain knowledge in a structure that is amenable to precise querying and reasoning. General purpose KBs such as Wikidata \citep{Vrandecic2012-aa} contain broad contextual knowledge, and are used for a wide variety of tasks, such as integrative analyses of otherwise disconnected data and enrichment of web applications (for example, a recipe website may want to dynamically query Wikidata to retrieve information about ingredients or country of origin). In the life sciences, KBs such as the Gene Ontology (GO) \citep{The_Gene_Ontology_Consortium2019-hv} and the Reactome biological pathway KB \citep{Fabregat2018-ml} contain extensive curated knowledge detailing cellular mechanisms that involve interacting gene products and molecules. These domain-specific KBs are used for tasks such as interpreting high-throughput experimental data. All KBs, whether general purpose or domain-specific, owe their existence to curation, often a concerted effort by human experts. 

However, the vast majority of human knowledge is communicated via natural language, with scientific knowledge communicated textually in journal abstracts and articles, which has historically been largely opaque to machines. The latest Natural Language Processing (NLP) techniques making use of Large Language Models (LLMs) have shown great promise in interpreting highly technical language, as demonstrated by orange their performance on question-answering benchmarks \citep{Ateia2023-ka}. These techniques have known limitations, such as being prone to hallucinations \citep{Ji2022-cq} (i.e., generating incorrect statements) and insensitivity to negations \citep{Ettinger2020-em}. Applications such as clinical decision support require precision and reliability not yet demonstrated by LMs of any size, though recent demonstrations offer promise \citep{Wang2020-pr,Khambete2021-uu,Luo2022-ed,Wachter2023-dj}.

If instead of passing the unfiltered results of LLM queries to users, we use LLMs to build KBs using NLP at the time of KB construction, then we can assist manual knowledge curation and acquisition while validating facts prior to insertion into the KB. NLP can assist KB construction at multiple stages. Literature triage aids selection of relevant texts to curate; Named Entity Recognition (NER) can identify textual spans mentioning relevant things or concepts such as genes or ingredients; grounding maps these spans to persistent identifiers in databases or ontologies; Relation Extraction (RE) connects named entities via predicates such as ‘causes’ into simple triple statements. Deep Learning methods such as autoregressive LMs \citep{Vaswani2017-np} have made considerable gains in all these areas. The first generation of these methods relied heavily on task-specific training data, but the latest generation of LLMs such as GPT-3 and GPT-4 are able to generalize and perform zero-shot or few-shot learning on these tasks by reframing them as prompt-completion tasks \citep{Brown2020-jf}.

\begin{figure}[t]
\centering
\includegraphics[width=\columnwidth]{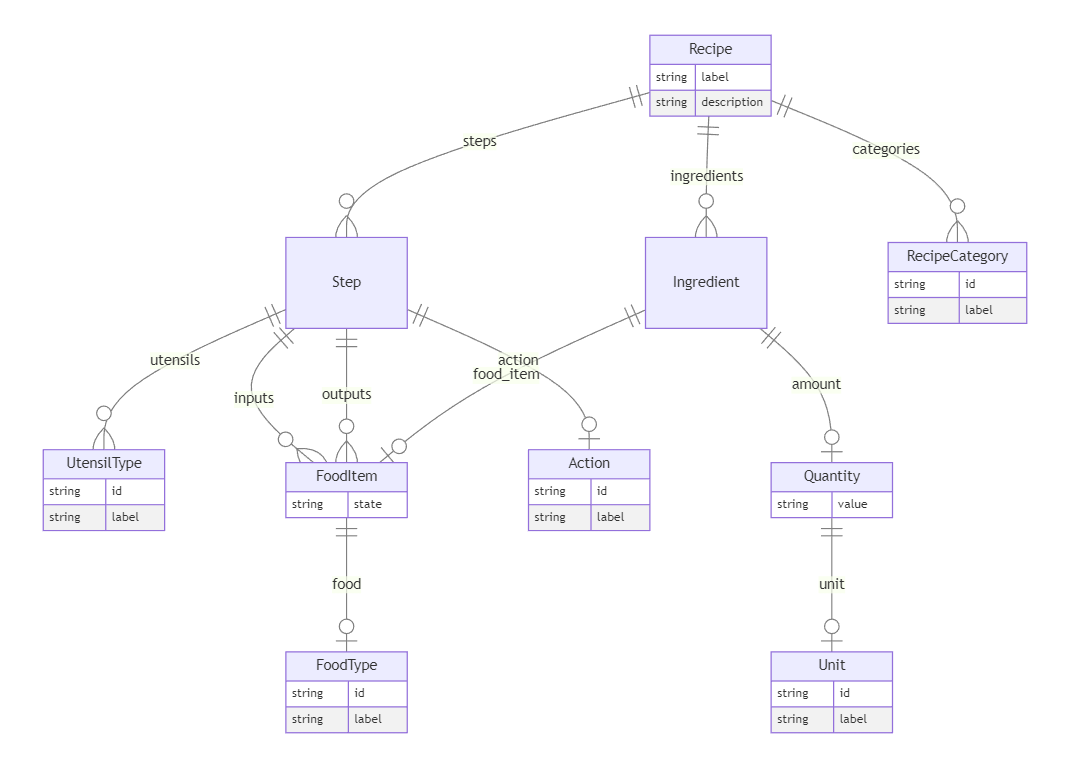}
\caption{Example schema. Boxes denote classes and arrows denote attributes whose range are classes (compound attributes). Crows feet above boxes denote multivalued attributes. Attributes whose ranges are primitives or value sets are shown within each box. Here, the top level container class “Recipe” is composed of a label, description, categories, steps, and ingredients. Steps and ingredients are further decomposed into food items, quantities, etc.}
\label{fig1}
\end{figure}

Most KBs are built upon detailed knowledge schemas which prove challenging to populate. Schemas describe the forms in which data should be structured within a domain. For example, a food recipe KB may break a recipe down into a sequence of dependent steps, where each step is a complex knowledge structure involving an action, utensils, and quantified inputs and outputs. Inputs and outputs might be a tuple of a food type plus a state (e.g. cooked) (Figure \ref{fig1}). Ontologies such as FOODON \citep{Wang2020-pr} may be used to provide identifiers for any named entities. Similarly, a biological pathway database might break down a cellular program into subprocesses and further into individual steps, each step involving actions, subcellular locations, and inputs and outputs with activation states and stoichiometry. Adapting existing pipelines to custom KB schemas requires considerable engineering.

\begin{figure}[ht]
\centering
\includegraphics[width=0.6\columnwidth]{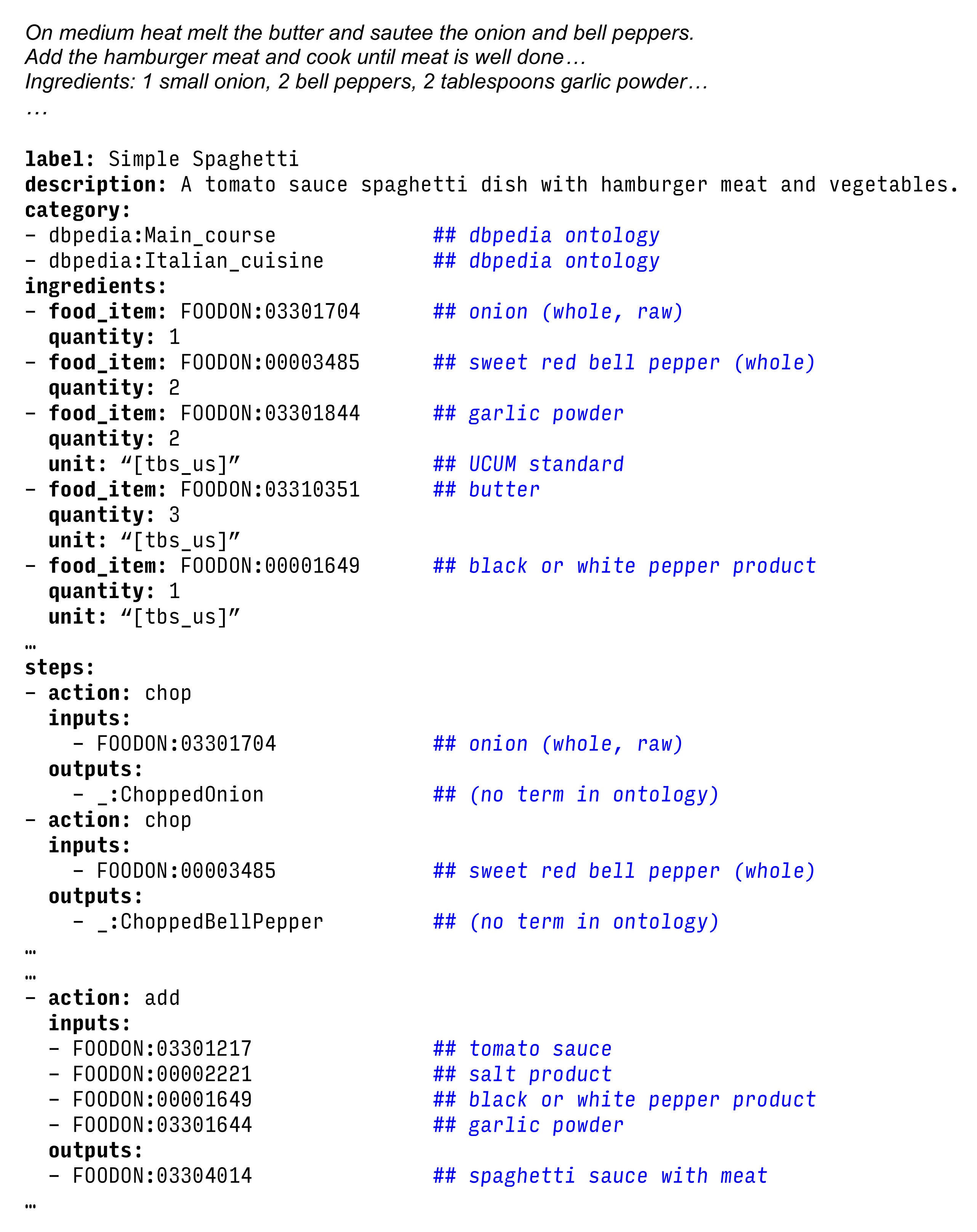}
\caption{Example of a portion of text to parse and a corresponding instantiation of the recipe schema from Figure \ref{fig1}, using YAML syntax. Input text is truncated for brevity; the full input is available at \url{https://github.com/monarch-initiative/ontogpt/blob/main/tests/input/cases/recipe-spaghetti.txt}. In each attribute-value pair, the attribute is shown in bold, followed by a colon and then the value or values. For multivalued attributes, each list element value is indicated with a hyphen at the beginning of the line. Terminal elements that are value sets from ontologies and standards such as FOODON \citep{Dooley2018-xn}, UCUM \citep{Schadow1999-ov}, and DBPedia \citep{Bizer2009-ts} are shown here with their human-readable labels in blue after the double-hash comment symbol. Dynamic elements are indicated via RDF blank node syntax (e.g. \texttt{\_:ChoppedOnion} does not correspond to a named entity and serves as a placeholder.}
\label{fig2}
\end{figure}

A schema provides a structure for data. For example, the recipe schema in Figure \ref{fig1} could be used in a recipe database, with each record instantiating the recipe class, with additional linked records instantiating contained classes, e.g. individual ingredients or steps. Figure \ref{fig2} shows an example of an instantiated schema class, rendered using YAML \citep{Ben-Kiki_undated-dz} syntax.

\begin{figure}[ht]
\centering
\includegraphics[width=0.5\columnwidth]{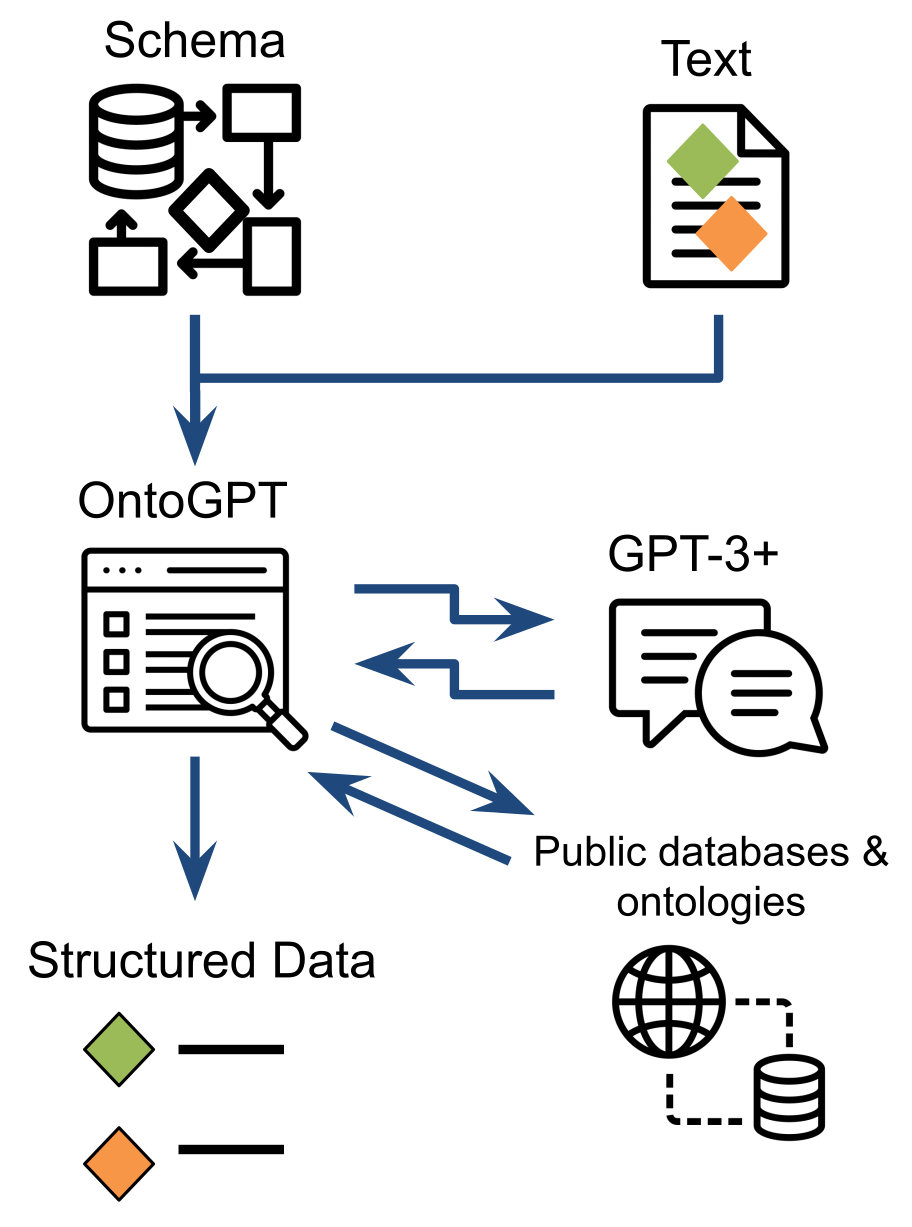}
\setlength{\belowcaptionskip}{-5pt}
\caption{Overview of the SPIRES approach. A knowledge schema and text containing instances defined in the schema are processed by OntoGPT, yielding a query for GPT-3 or newer, accessed through the OpenAI API. OntoGPT parses the result, grounding extracted instances with specific entries and terms retrieved from queries of databases and ontologies where possible. The final product is a set of structured data (instances and relationship) in the shapes defined by the schema. Icons by user Khoirin from the Noun Project (\url{https://thenounproject.com/besticon/}).
}
\label{fig3}
\end{figure}

Here we present Structured Prompt Interrogation and Recursive Extraction of Semantics (SPIRES), an automated approach for population of custom schemas and ontology models. The objective of SPIRES is to generate an instance (i.e., an object) from a text, where that instance has a collection of attribute-value associations. Each value is either a primitive (e.g. string, number, or identifier) or another inlined instance (Figure \ref{fig2}). SPIRES integrates the flexibility of LLMs with the reliability of publicly-available databases and ontologies (Figure \ref{fig3}). This strategy allows SPIRES to fill out schemas with linked data while bypassing a need for training examples. A major advantage of SPIRES over more traditional RE is its ability to populate schemas that exhibit nesting, in which complex classes may have attributes whose ranges are themselves complex classes. SPIRES also makes use of a flexible grounding approach that can leverage over a thousand ontologies in the OntoPortal Alliance \citep{Graybeal2019-nz}, as well as biomedical lexical grounders such as Gilda \citep{Gyori2022-xx} and OGER \citep{Furrer2019-tt}. This grounding method offers far more consistent mapping to unique identifiers than hallucination-prone LLM querying alone.

\section{System and Methods}\label{sec2}

In SPIRES, A knowledge \textit{schema} is a structure for constraining the shape of instances for a given domain. A schema is a collection of \textit{classes} or templates, each of which can be instantiated by instances. Each class has a collection of \textit{attribute constraints}, which control the attribute-value pairs that can be associated with each instance. The \textit{range} of an attribute specifies the allowed value or values. A range can be either (1) a primitive type such as a string or number; (2) a class; or (3) an \textit{enumeration} of permissible value tokens (e.g., an enumeration of days of the week may include "Monday", "Tuesday", and so on). Attributes also have \textit{cardinality}, specifying the minimum and maximum number of values each instance can take. Additionally, each schema element can have arbitrary metadata associated with it. 

Formally, a schema \textit{S} consists of \textit{n} classes:

\begin{equation}
Classes\left ( S \right ) = \left \{ c_1, \cdots , c_n \right \}
\end{equation}

Classes correspond to the kinds of entities present in a database (e.g. in a recipe database, this would include recipes, as well as ingredients and steps; see example in Figure \ref{fig1}).

Each class $c_i$ has an ordered list of attributes:

\begin{equation}
Attributes\left ( c_i \right ) = \left \{ c_ia_1, \cdots , c_ia_m \right \}
\end{equation}

Instances of $c_i$ may have \textit{values} specified for each of these attributes. An attribute \textit{a} can have associated properties:
\begin{itemize}
    \item $Name(a)$ : the name of the attribute; for example, “summary” or “steps”.
    \item $Multivalued(a)$ = $\{True, False\}$, indicating whether the value of a is a list, or single-valued. A recipe might have a single-valued attribute for the name of the recipe, and a multivalued attribute for the steps.
    \item $Identifier(a)$ = $\{True, False\}$, indicating whether \textit{a} is a persistent identifier for instances, such as the FOODON identifiers in Figure \ref{fig2}.
    \item $Prompt(a)$ = string, which is a user-specified custom prompt for that attribute.
    \item $Range(a)$: the allowable values for this attribute; this can be a class \textit{c} in \textit{S}, or a primitive type such as string or number, or a value set (see below). In Figure \ref{fig1}, the range of the \textit{ingredients} attribute is Ingredient, and the range of the \textit{id} attribute is a string.
    \item $ValueSets(c)$: a list of atomic values from which values of \textit{a} can be drawn, where a value set is either an extensional list (fixed/static) or intensional (specified by ontology query). For example, a value set for a food element in an ingredient may be drawn from the food branch of the Food Ontology.
    \item $Inlined(a)$ = $\{True, False\}$, indicating, when the range is a class, if the object should be nested/embedded, or passed by reference.
\end{itemize}

Additionally, a class \textit{c} can include a set of identifier constraints:
\begin{equation}
IDSpaces\left ( c_i \right ) = \left \{ prefix_i, \cdots , prefix \right \}
\end{equation}

The constraint set is a list of strings that are the allowable prefixes that the identifier can take--for example, “WIKIDATA”, “MESH”, “GO”, or “FOODON”. The prefixes should come from a standard prefix registry such as BioRegistry \citep{Hoyt2022-vr} to ensure consistency across schemas and projects; SPIRES expects upper-case prefixes.

\subsection{Evaluation of Entity Grounding}\label{subsec2-1}

To determine the extent to which SPIRES improves entity grounding relative to prompting alone, we queried two GPT models with sets of ontology term labels with and without our grounding. We selected 100 terms at random from each of three ontologies: the Gene Ontology (GO), the Mouse Developmental Anatomy Ontology (EMAPA), and the MONDO Disease Ontology. The 16k GPT-3.5-turbo (gpt-3.5-turbo-16k) and the newly available GPT-4-turbo (gpt-4-1106-preview) models were each queried with the full term list in a single prompt each along with text requesting corresponding identifiers from the specified ontology (or, for SPIRES, a structured query based on a minimal schema). A match was considered successful for each pair of identifier and label in which the label text was parsed as a single entity, remained unchanged in the output, and matched to the correct identifier. The full evaluation and results are available in a code notebook online\footnote{\url{https://github.com/monarch-initiative/ontogpt-experiments/blob/main/experiments/ground_compare/Comparing_Grounding.ipynb}}.

\subsection{Evaluation Against Chemical Disease Relation Task}\label{subsec2-2}

We evaluated SPIRES on the Biocreative Chemical-Disease-Relation task \citep{Li2016-yx}. We used all 500 abstracts of the BC5CDR test set and evaluated against the set of 1066 chemical-induces-disease (CID) triples. For each triple, the predicate is fixed, and the subject and object are always identifiers drawn from the Medical Subject Headings (MeSH) vocabulary \citep{Lipscomb2000-zv}. Grounding was performed using multiple ontologies beyond MeSH, including three resources for chemical and drug information: Chemical Entities of Biological Interest (ChEBI) \citep{Hastings2016-fl}, DrugBank \citep{Wishart2018-xr}, and MedDRA \citep{Brown1999-zq} (See Table \ref{tableS1} for a full list of external resources used for grounding). We used the Translator NodeNormalizer \citep{Fecho2022-nk} to normalize these to MeSH IDs to permit comparison with the test set. No fine tuning was performed. The training set was used to enhance our mappings of named entity spans to MeSH identifiers; after building this lexicon, the training set was discarded.

We provided SPIRES with a model of chemical to disease (CTD) associations based on the Biolink Model \citep{Unni2022-qy}. Biolink extends the simple triple model of associations to include qualifiers on the predicate, subject, and object. Subject and object qualifier information was discarded in this evaluation as extracting these details was not tested for in the original CDR benchmark. Statements with predicate qualifiers of “NOT” were discarded. We configured value sets for MeSH Disease and Chemical entries manually (see the full list of identifiers used to define these sets in Table \ref{tableS2}). NER of chemical and disease entities was also evaluated based on ability to identify a corresponding MeSH. We compared two pre-processing approaches: a "chunking" approach in which input documents were processed as separate subsegments (essentially a sliding window approach) and a "no chunking" approach in which the entirety of the test corpus document title and abstract was passed in a prompt. Two OpenAI models were used in these comparisons: gpt-3.5-turbo and gpt-4.

\section{Algorithm}\label{sec3}

\begin{figure}[ht]
\centering
\includegraphics[width=\columnwidth]{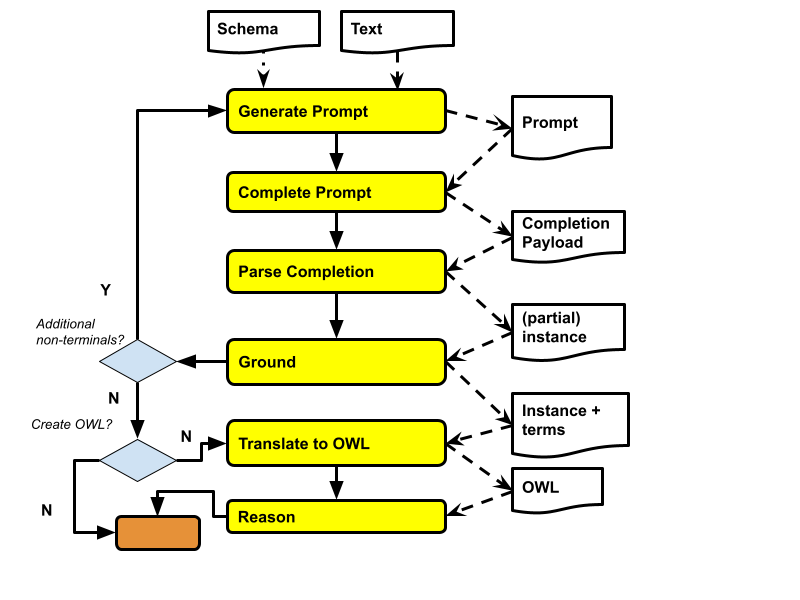}
\setlength{\belowcaptionskip}{-10pt}
\caption{Flowchart depicting the SPIRES algorithm.
}
\label{fig4}
\end{figure}

The SPIRES extraction procedure takes as input (1) a schema \textit{S}, (2) an entry point class \textit{C}, and (3) a text \textit{T} (Figure \ref{fig4}, top). It returns a structured instance \textit{i} conforming to \textit{S}, making use of a large language model (LLM) that allows prompt completion, such as GPT-3 and its more recent versions. The procedure is detailed below: \\

\noindent${SPIRES}(S, C, T)$:
\begin{enumerate}
\item Generate the prompt: $p = GeneratePrompt(S, C, T)$
\item Perform prompt completion: $r = CompletePrompt(p)$
\item Parse results and recurse over nested structures: \\ $iu = ParseCompletion(r, S, C)$
\item Ground results using ontologies: $i = Ground(iu, S, C)$
\item (optional) translation to OWL: $ont=TranslateToOWL(i)$
\end{enumerate}

\subsection{Step 1: Generate Prompt}\label{subsec3-1}

SPIRES first generates text for a prompt (Figure \ref{fig4}, \textit{Generate Prompt}) to be provided to the LLM:

\begin{equation}
GeneratePrompt\left( S, C, T \right) = Instructions\left(\right) 
+ AttributeTemplate\left( S, C, T \right) + TextIntro\left(\right) + T + Break\left(\right)
\end{equation}

Here, the \textit{Instructions} function returns a piece of text such as “From the text below, extract the following entities in the following format”.

The \textit{AttributeTemplate} function generates a pseudo-YAML structure that is a template for results. For each \textit{a} in $Attributes(C)$, we write:
\begin{equation}
Name\left( a \right) + ":" + Prompt\left( a \right) + "\backslash n"
\end{equation}
If Prompt is undefined for attribute \textit{a}, then it is automatically generated from the name. If $Multivalued(a)$ is True, then the text is preceded with “A semicolon-separated list”.

The \textit{TextIntro} function introduces a break between the template and the input text and is a fixed string “Text:”. The \textit{Break} function is also a fixed string that serves to demarcate the end of the text and is a sequence of three break characters, e.g. “===”. As an example, when calling this function when \textit{S}=RecipeSchema, \textit{C}=Ingredient, and \textit{T}=“garlic powder (2 tablespoons)”, the following prompt would be generated:\\

\begin{samepage}
\ttfamily

\noindent Split the following piece of text into fields in the\\ following format:
\\\\
food\_item: <the food item> \\
amount: <the quantity of the ingredient>
\\\\
Text: \\
garlic powder (2 tablespoons) \\

\noindent===
\normalfont
\end{samepage}

Note that typical input texts will be larger, except when the function is called recursively.

\subsection{Step 2: Complete the Prompt}\label{subsec3-2}

The generated prompt is provided to the LLM using a completion API (Figure \ref{fig4}, \textit{Complete Prompt}). The nature of the prompt can be adapted for different language models; the OntoGPT implementation defaults to the orange GPT-3.5-turbo model \citep{openai_undated-tt} but is compatible with any model capable of delivering a payload orange that conforms to a prompt-specified structure. The intended completion results are a pseudo-YAML structure conforming to the specified template. For example, when passing the example prompt in Step 1, the return payload may be the following text:

\ttfamily
\noindent
food\_item: garlic powder \\
amount: 2 tablespoons

\normalfont

\subsection{Step 3: Completion Result Parsing and Recursive Extraction}\label{subsec3-3}

The $ParseCompletion(r, S, C)$ function returns a pre-grounded instance object \textit{i} partially conforming to \textit{C}. This step consists of two sub-steps: (1) parsing the pseudo-YAML; (2) recursively calling SPIRES on any inlined attributes. For the parsing step (Figure \ref{fig4}, \textit{Parse Completion}), the completion provided by the LLM is not guaranteed to be strict YAML or even conform directly to the specified template, so a heuristic approach is used. The response is separated by newlines into a list. Each line is split on the first instance of a “:”; the part before is matched against the attribute name, and the part after is the value, which is parsed as detailed below. Attribute matching is case-insensitive. All whitespace is normalized to underscores.

Each value \textit{v} is parsed according to the range and cardinality of the matched attribute \textit{a}, populating each attribute \textit{a} of \textit{i}:
\begin{equation}
i[a] = ParseValue(v)
\end{equation}
If \textit{a} is \textit{multivalued}, then \textit{v} is first split according to a delimiter (default “;”), and the rules below are applied on each token; otherwise the rules below are applied directly.

\noindent Rule 1: If the range is a primitive data type (i.e. string, number, or boolean) then the value is returned as-is. 

\noindent Rule 2: If the range of the attribute is a class, and the attribute is non-inlined (i.e. a reference) or an enumeration, then the value will be grounded, as specified in Step 4 below.

\noindent Rule 3: if the range of the attribute is an inlined class, then SPIRES is called recursively:
\begin{equation}
SPIRES(S, Range(a), v)
\end{equation}
This proceeds until a non-inlined class is reached. For example, given the example payload from the previous step, the attribute \textit{food item} is a reference to an ontology class, so the value “garlic powder” is grounded using the grounding procedure (Step 4). The attribute amount is a reference to an inlined class \textit{Quantity}, so this will be recursively parsed by calling $GeneratePrompt(RecipeSchema, Quantity, \texttt{"2 tablespoons"})$.

\subsection{Step 4: Grounding and Normalization}\label{subsec3-4}

All leaf nodes of the instance tree that correspond to named entities are grounded, i.e., mapped to an identifier in an existing vocabulary, ontology, or database (Figure \ref{fig4}, \textit{Ground}). Classes representing named entities can each be annotated with one or more vocabularies. Each vocabulary is identified by a unique prefix. For example, in Figure \ref{fig1}, the FoodItem class could be annotated with both FOODON and Wikidata, indicating that grounding on labels can be performed using these vocabularies. Grounding on the string “garlic powder” may then yield FOODON:03301844 when the BioPortal \citep{Whetzel2011-zm} or AgroPortal annotator \citep{Jonquet2018-bv} is used, and WIKIDATA:Q10716334 when a Wikidata normalizer is used. The final results are normalized via validation against identifier constraints for the class. If $IDSpaces(c)$ is set, then the prefix of the identifier is checked against the list of valid prefixes. If $ValueSets(c)$ is set, then the value returned must be present in the value set.

\subsection{Step 5: Translation to OWL and Reasoning}\label{subsec2-2-5}

Step 4 produces an instance tree that can be directly represented in JSON or YAML syntax (both of which allow for arbitrary nesting of objects). For some KBs, this is sufficient. Further conversion to an ontological representation in OWL (Figure \ref{fig4}, \textit{Translate to OWL}), and additional reasoning steps, then support checking for consistency and population of missing axioms. There are multiple methods for translating to OWL, including ROBOT templates \citep{Jackson2019-ae}, DOSDPs \citep{Osumi-Sutherland2017-an}, and OTTR \citep{Kindermann_undated-mh}.

\section{Implementation}\label{sec4}

We provide an implementation of SPIRES in Python as part of the OntoGPT Python package\footnote{\url{https://github.com/monarch-initiative/ontogpt}}, which provides both a command line interface (CLI) and a simple web application (Supplementary Figure \ref{figS1}). SPIRES uses LinkML \citep{Moxon2021-ds} as its Knowledge Schema language. This allows for a full representation of the necessary schema elements while incorporating LinkML's powerful mechanism for specifying static and dynamic value sets. For example, a value set can be constructed as a declarative query of the form “include branches \textit{A}, \textit{B} and \textit{C} from ontology $O_1$, excluding sub-branch \textit{D}, and include all of ontology $O_2$”. The LinkML framework also supports converting schemas to LinkML from forms such as SHACL \citep{Pareti2022-eq}, JSON-Schema \citep{jsonschema_undated-zx}, or SQL Data Definition Language, allowing their use with SPIRES.

SPIRES performs grounding and normalization with the Ontology Access Kit library (OAKlib) \citep{Mungall2023-yj}, which provides interfaces for multiple annotation tools (i.e., those providing links to external vocabularies and ontologies), including the Gilda entity normalization tool \citep{Gyori2022-xx}, the BioPortal annotator \citep{Jonquet2009-is}, and the Ontology Lookup Service \citep{Jupp_undated-az}. For identifier normalization a number of services can be used, including OntoPortal mappings, with the default being the NCATS Biomedical Translator Node Normalizer \citep{Fecho2022-nk}.

The results of extraction can optionally be further processed using LinkML-OWL \citep{Mungall2022-we}, which generates an OWL representation of instance data using mappings specified in a LinkML schema. This OWL file can be used as an input to ROBOT \citep{Jackson2019-ae} to run OWL reasoning to check for logical inconsistencies and perform automated classification.

\subsection{Standard Templates for Multiple Applications}\label{subsec4-1}

\begin{table}
\caption{Pre-made schemas. Example use cases are included but are not comprehensive. Note the CTD schema is deliberately restricted to only use the MESH vocabulary for purposes of evaluation. Identifiers refers to all ontologies, value sets, and other unique term sets incorporated in a given schema.}
\begin{tabular}{p{2.5cm}p{4.5cm}p{2cm}p{5.5cm}}
\toprule
Schema & Use Case & Identifiers & Text inputs \\
\midrule
Food Recipes & Enforcing consistent structure on stepwise processes                  & FOODON, UO              & Unstructured and semi-structured recipes                                                        \\
Drug mechanisms & Integrating drug descriptions            & MONDO, CHEBI, MESH            & Mechanism of Action (MOA) descriptions                                                    \\
Chemical-disease interactions & Assembling knowledge graphs of chemical-impacted phenotypes      & MESH                          & Abstracts describing effects of chemicals on conditions                                                      \\
Metagenomic Samples & Standardizing metadata for metagenomics       & ENVO                          & Descriptions of environmental samples     \\
Mendelian Diseases & Extracting disease relationships from literature          & MONDO, HPO              & Case studies or descriptions of Mendelian diseases \\ 
\bottomrule
\end{tabular}
\label{table1}
\end{table}

The SPIRES implementation comes with a growing collection of ready-made schemas for multiple applications. These are primarily life-science focused, for example, deriving a pathway from a Mechanism of Action description in a database such as DrugBank. We also include a schema for food recipes to demonstrate general applicability in domains beyond the environmental and life sciences. Table \ref{table1} lists a selection of the pre-made schemas.

\subsection{Extraction of Recipe Ontologies from Websites}\label{subsec4-2}

To demonstrate the full functionality of OntoGPT we created a pipeline for extracting recipes from websites and generating an OWL ontology from the combined outputs. Recipes are extracted using the recipe-scrapers Python module\footnote{\url{https://github.com/hhursev/recipe-scrapers}}. The pipeline takes the output of scraping, concatenates the results into a text, then feeds this to OntoGPT using the recipe template. We use LinkML-OWL to map the recipe template to OWL axioms, such that each recipe is represented as a class defined by its ingredients and its steps. We use ROBOT to extract the relevant parts of the FOODON ontology, and merge this with the extraction results, combined with a manually coded simple recipe classification with defined classes for groupings such as “Meat Recipe” and “Wheat Based Recipe”. We use the Elk reasoner \citep{Kazakov2015-nl} to classify the results. The results of this process are highlighted in Figure \ref{figS3}.

\subsection{Entity Grounding}\label{subsec4-3}

Grounding entities with ontology terms is part of the core functionality of SPIRES and its value is well demonstrated in a direct comparison with the straightforward approach of directly querying an LLM with term descriptions. If we request the GO term for "integrase activity" we expect the response to include GO:0008907, for example. Of 100 GO terms chosen at random, SPIRES returned the correct identifiers for 98 when using GPT-3.5-turbo and 97 with GPT-4-turbo. Without SPIRES, GPT-3.5-turbo returned just 3 correct identifiers. Though it yielded 100 putative matches, few included correct GO identifiers. This "mass hallucination" may be an artifact of prompting with terms lacking surrounding context. Even so, it may be challenging to determine how much context is sufficient to improve grounding. GPT-4-turbo demonstrated a different challenge by consistently refusing to retrieve identifiers, returning responses such as "As an AI developed before 2023, I do not have real-time access to databases...". For the EMAPA mouse anatomy ontology, SPIRES returned correct identifiers for all 100 term descriptions, while GPT-3.5-turbo repeatedly provided identifiers from the EHDAA2 human anatomy ontology instead. GPT-4-turbo refused to ground EMAPA terms as it had with GO. MONDO terms posed some surprising difficulty: SPIRES with GPT-3.5-turbo correctly returned 97 of 100 identifiers but SPIRES with GPT-4-turbo returned just 18 correct matches. In some cases, this may have been due to incorrectly parsing entities (e.g., parsing "UV-induced skin damage, susceptibility to" as "skin damage"). As with GO, prompting without SPIRES only returned one correct identifier at most from both GPT-3.5-turbo and GPT-4-turbo.

\subsection{Evaluation on BioCreative Chemical Disease Relation Task}\label{subsec4-4}

We evaluated SPIRES on the BioCreative Chemical-Disease-Relation (BC5CDR) task corpus. To demonstrate the zero-shot learning approach, we did not perform any fine tuning using the training set. The training set was used to enhance our mappings of named entity spans to MeSH identifiers and was then discarded. For our CTD schema (see Figure \ref{figS2}), we follow the Biolink Model \citep{Unni2022-qy} which extends the simple triple model of associations to include qualifiers on the predicate, subject, and object. This yields finer-grained predictions; for example, SPIRES correctly parses the statements in Table \ref{table2}. In these cases, SPIRES grounds the drug entity Cromakalim to its corresponding MeSH identifier and extracts its relationship with vasodilation along with a qualifier noting the observation is specific to "large and small coronary vessels", an anatomical entity worthy of further grounding (though this was not explored within the original BC5CDR task). Similarly, the correctly extracted relationship between lithium and hypercalcemia includes the qualifier that the observation pertains to chronic lithium exposure.

\begin{table}[t]
\caption{Extracted relation examples. All predicates are 'INDUCES'. Sources are PubMed identifiers (PMIDs). PMID 2160002 is “Vasodilation of large and small coronary vessels and hypotension induced by cromakalim and pinacidil” \citep{Giudicelli1990-qb}. PMID 19154241 is a case report on lithium therapy \citep{Rizwan2009-xo}. PMID 10327032 is a study of hyperammonemic encephalopathy risks in cancer patients \citep{Liaw1999-ra}.}
\begin{tabular}{p{1.5cm}p{3cm}p{1.5cm}lp{3cm}p{3cm}}
\toprule
Source & Subject & Sub. qual. & Predicate & Object & Object qual. \\
\midrule
2160002 &
  MESH:D019806  \mbox{Cromakalim} &
  - &
  INDUCES &
  MESH:D014664  \mbox{Vasodilation} &
  large and small \mbox{coronary} vessels \\
2160002  &
  MESH:D020110  Pinacidil &
  - &
  INDUCES &
  MESH:D014664 \mbox{Vasodilation} &
  large and small \mbox{coronary} vessels \\
19154241  &
  MESH:D008094   Lithium &
  Chronic &
  INDUCES &
  MESH:D006934 \mbox{Hypercalcemia} &
  - \\
10327032  &
  MESH:D005472 \mbox{Fluorouracil} &
  - &
  INDUCES &
  MESH:D001927 Brain Diseases &
  Transient \\
\bottomrule
\end{tabular}
\label{table2}
\end{table}

When evaluating, we discard subject and object qualifier information, as this is not tested for in the original CDR benchmark. If the predicate qualifier is “NOT” then we discard the whole statement. Note that in the examples in Table \ref{table2}, even though we evaluated the first two statements to be a correct interpretation of the abstract, they were counted as false negatives; the corresponding triple was not in the test set, presumably an error of omission.

For SPIRES, we saw initially encouraging results on the BC5CDR task with chunking and GPT-3.5-turbo: we observed an F-score of 41.16, precision of 0.43, and recall of 0.39. Using the "no chunking" approach (i.e., no preprocessing of the test document) yielded an F-score of 36.64 (precision 0.63, recall 0.26) with GPT-3.5-turbo and an F-score of 43.80 (precision 0.69, recall 0.32). For NER results alone (i.e., correct grounding against MeSH for chemical and disease entities), see Table \ref{tableS3}.

These results place SPIRES just below the average of all 18 teams that participated in the original CDR challenge. We assume all 18 teams used the full training set, whereas with SPIRES there was no task-specific training or fine tuning. For comparison, Luo et al. report an F-score of 44.98 on BC5CDR with their biomedical domain-specific, trained-from-scratch BioGPT model \citep{Luo2022-ed}. We note that the best-scoring relation extraction results from the CDR task achieved an impressive score of 0.57, though with a model trained on a large and carefully engineered set of training examples \citep{Xu2015-kf}. SPIRES bypasses this step but may see further improvement with fine-tuned and/or domain-specific LLMs.

\section{Discussion}\label{sec5}

\subsection{Comparable Methods}\label{subsec5-1}

SPIRES is a well-developed and generally model-agnostic approach for information extraction designed with structured schemas and standardized ontologies in mind. Some recent efforts have made great strides in leveraging the first type of resource, i.e., they address the task of aligning extracted information with pre-defined data models. The fine-tuned GPT-3-based approach described by Dunn and Dagdelen et al. employs engineered schemas to extract structured relationships from unstructured text in materials chemistry \citep{Dunn2022-js}. The authors of the LLMs4OL approach also explored application of LLMs to information extraction, but concluded that the models are not yet sufficiently flexible for ontology-driven needs \citep{Babaei_Giglou2023-ym}. We also consider the task of ontology alignment to be related to our efforts; we have found that LLMs can noticeably improve accuracy in ontology alignment \citep{Matentzoglu2023-id} and the development of general frameworks such as Agent-OM \citep{Qiang2023-lv} may further improve the grounding inherent to information extraction.

\subsection{Choosing a Model}\label{subsec5-2}

OntoGPT currently supports both select open LLMs and the OpenAI API. Running OntoGPT across a large corpus with OpenAI models may be prohibitively expensive for some users. Additionally, the use of this API involves closed models with inscrutable training data, which may be plagued by biases \cite{Bender2021}. Though our experiments here generally concern GPT-3 and 4, the rapid pace of model development will ensure access to progressively more capable (and ideally, more transparent) language models. Smaller LMs such as LLaMA have been shown to outperform models ten times their size \citep{touvron2023}, and it is possible to fine-tune these into instruction following models \citep{Zhang2023}. LLMs based on LLaMA2 and adapted for biomedical language, including BioMedGPT-LM \cite{Luo2023-cw} and Radiology-Llama2 \cite{Liu2023-no}, may complement the grounding provided through SPIRES.

\subsection{Reliability and Hallucinations}\label{subsec5-3}

A common problem with LLMs is hallucination of results (producing factually invalid statements that are not consistent with the input text) \citep{Ji2022-cq,Bender2021}. We crafted prompts to limit hallucination, asking only for the LM to extract what was found in the text, and keeping default low-creativity settings. On examination we found that hallucinations were generally infrequent, with most false positives and negatives attributable to incorrect relation extraction. It is worth noting that LLM interfaces designed for direct function calling may duplicate some of the data structure enforcement afforded by SPIRES but do not alleviate the issue of hallucination: a model may still improperly associate real or fictional ontology identifiers with extracted entities when queried without aid of our approach.

Some text generation may yield technically correct results. For example, one result extracted from the title “Increased frequency and severity of angio-oedema related to long-term therapy with angiotensin-converting enzyme inhibitor in two patients”, yielded “Lisinopril INDUCES angio-oedema”. Lisinopril is in fact a subtype of ACE inhibitor, and the extracted association is supported by other literature. However, this more precise statement is not the one that is in the original text. Presumably the LM is substituting the class of drug with a specific member here, but it is unclear why it does it on this occasion. Until there are better methods to control this hallucination and explain justifications for statements in terms of the text and prior knowledge, results from LMs should be carefully validated before being entered into KBs.

SPIRES is a new approach to information extraction that leverages recent advances in large language models to populate complex knowledge schemas from unstructured text. It uses zero-shot learning to identify and extract relevant information from query text, which is then normalized and grounded using existing ontologies and vocabularies. SPIRES requires no model tuning or training data. The approach is customizable, flexible, and can be used to populate knowledge schemas across varied domains. We envision SPIRES being used not in isolation, but rather in synergistic strategies combining human expertise, linguistic pattern recognition, deep learning and classical deductive reasoning approaches. SPIRES is one component of a growing toolkit of methods for transforming noisy, heterogeneous information into actionable knowledge.

\section{Competing interests}
No competing interest is declared.

\section{Acknowledgements}\label{acknowledgements}

\subsection{Funding}\label{funding}
This work was supported by the National Institutes of Health National Human Genome Research Institute [RM1 HG010860]; National Institutes of Health Office of the Director [R24 OD011883]; and the Director,
Office of Science, Office of Basic Energy Sciences, of the US Department of Energy [DE-AC0205CH11231 to J.H.C., H.H., N.L.H., M.J., S.M., J.T.R, and C.J.M.]. We also gratefully acknowledge Bosch Research for their support of this research project.

\bibliographystyle{unsrtnat}
\bibliography{reference}

\clearpage
\setcounter{page}{1}
\setcounter{table}{0}
\renewcommand{\thetable}{S\arabic{table}}
\setcounter{figure}{0}
\renewcommand{\thefigure}{S\arabic{figure}}
\section{Supplementary Data}\label{supplement}

\begin{figure*}[h]

\centering
\includegraphics[width=468pt]{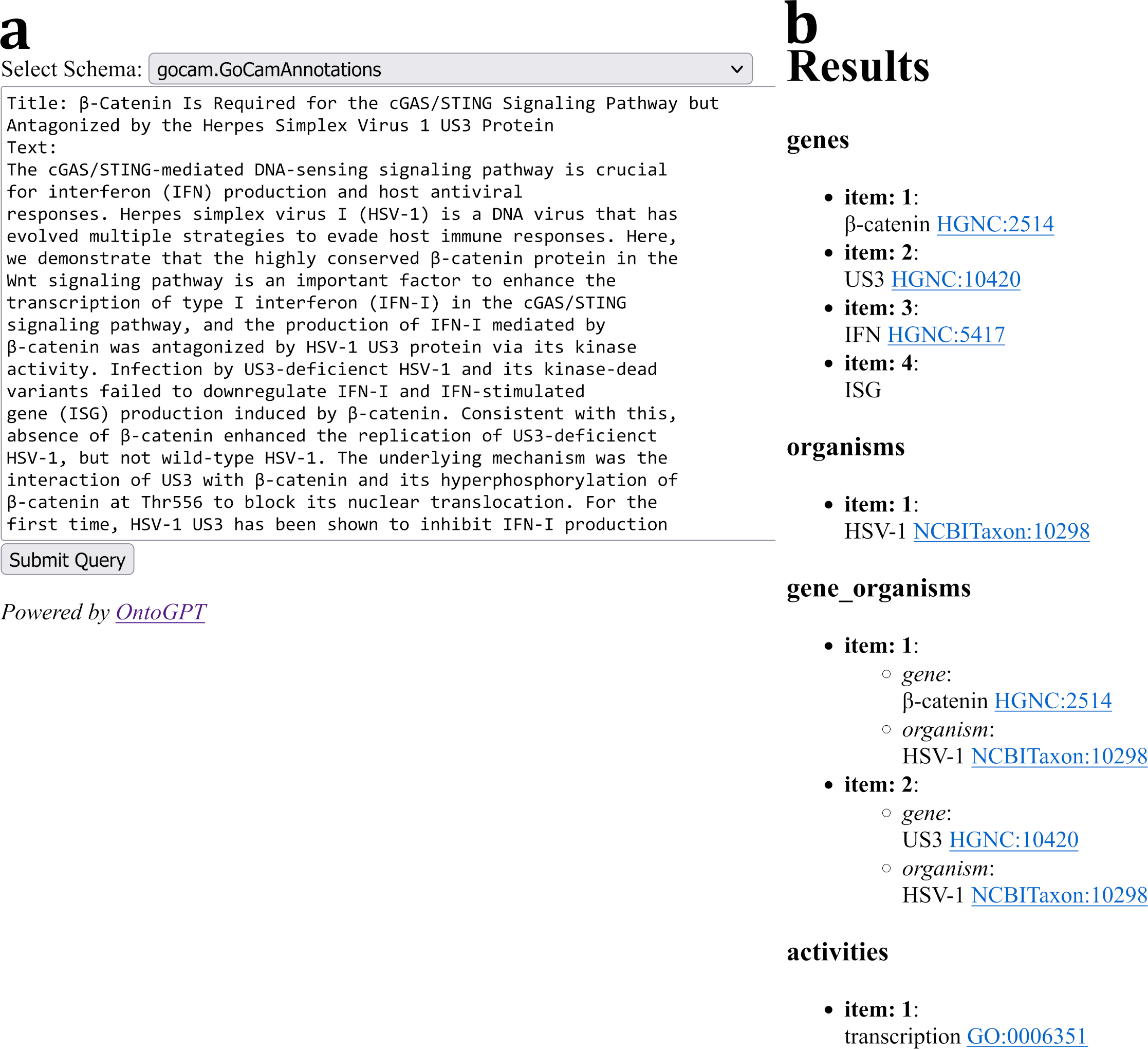}
\caption{Screenshot of web-ontogpt. (a) Form entry page, allowing selection of schema, plus input text. (b) Sample of results as structured object rendered as nested HTML. Note that both input text and results are truncated for brevity.
}
\label{figS1}

\end{figure*}

\begin{figure*}[ht]
\centering
\includegraphics[width=468pt]{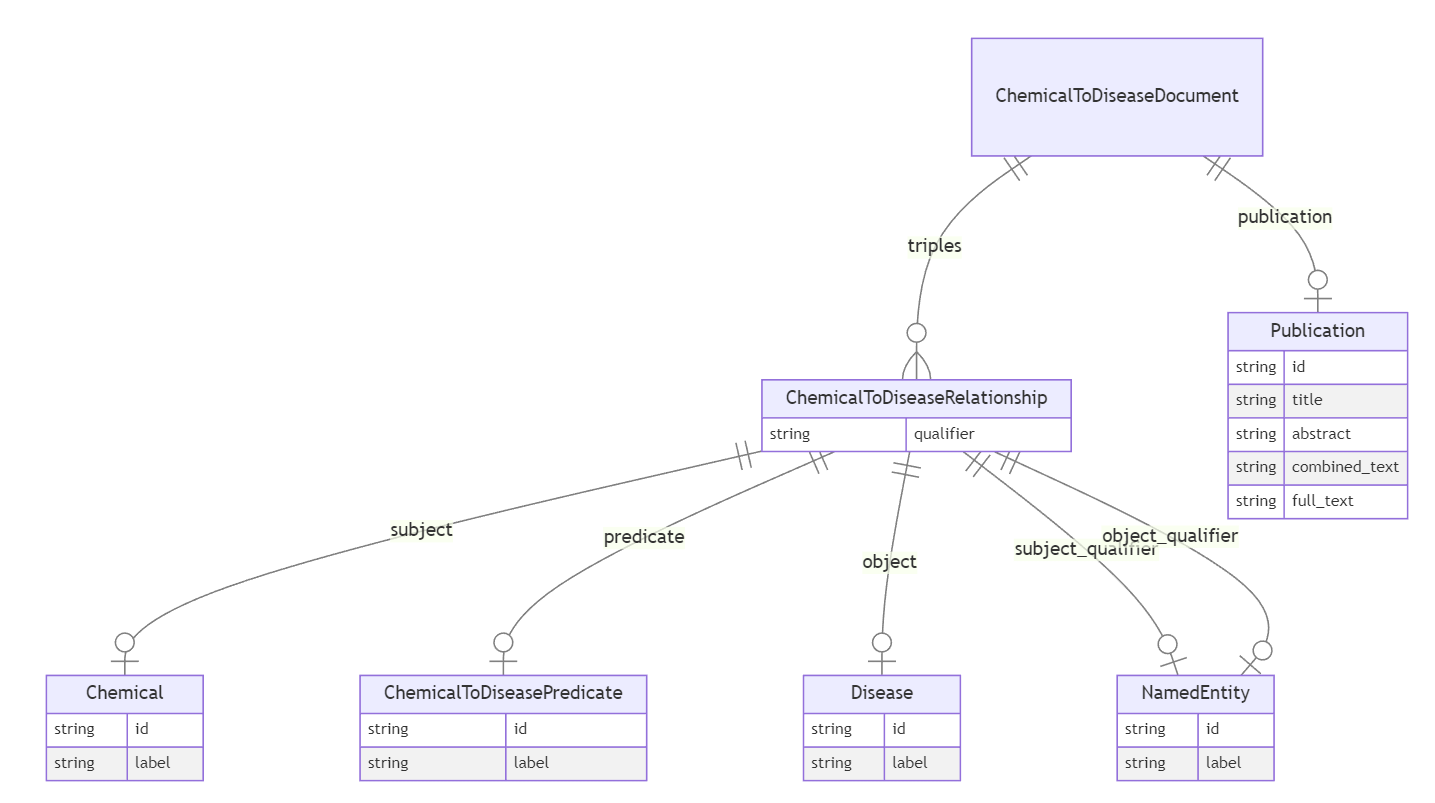}
\caption{Chemical to Disease (CTD) schema (available from \url{https://w3id.org/ontogpt/ctd}). 
}
\label{figS2}
\end{figure*}

\begin{figure*}[ht]
\centering
\includegraphics[width=\columnwidth]{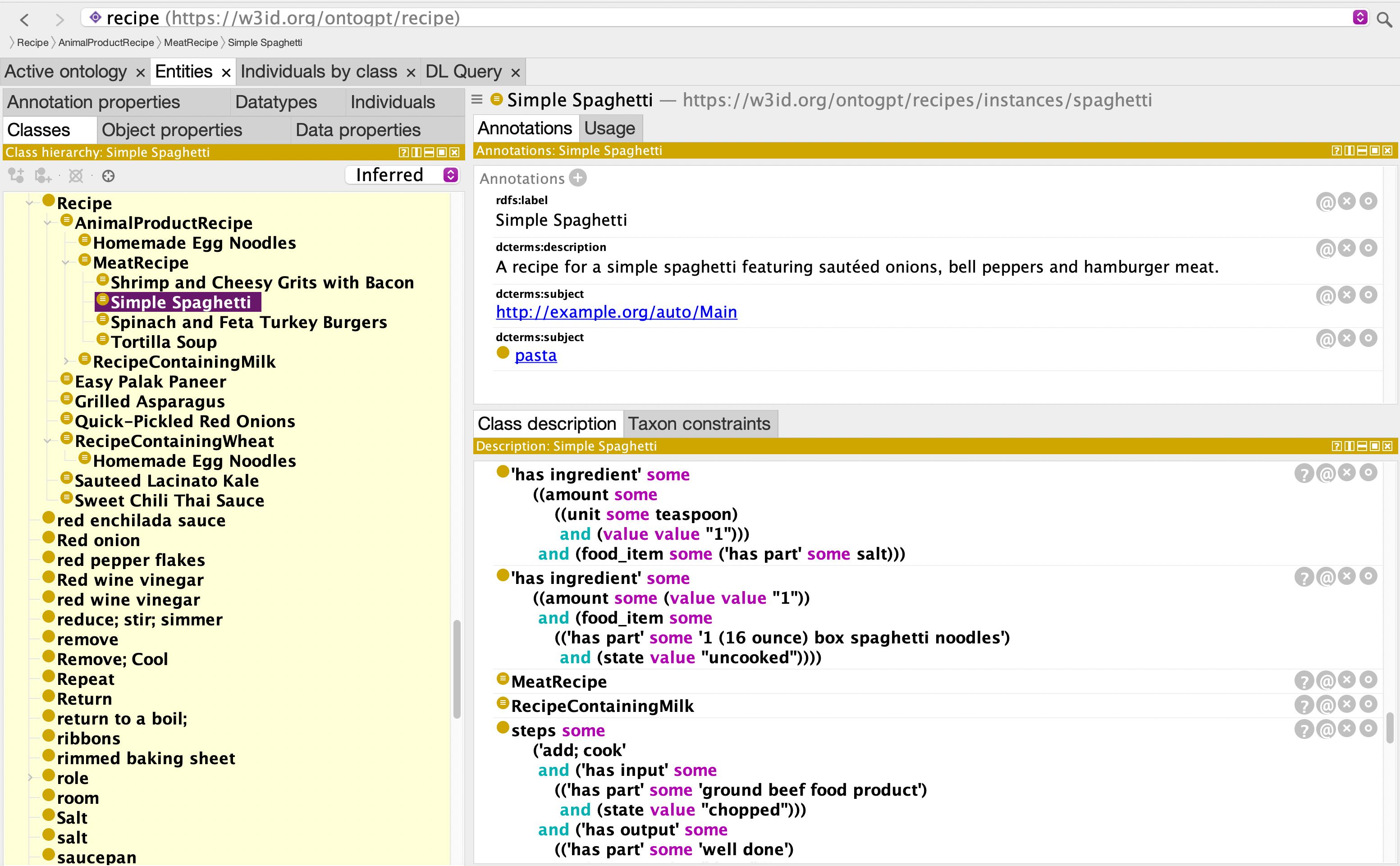}
\caption{Screenshot of extracted recipes in a merged OWL file from the Protege ontology editor \citep{Musen2015-bj}. The "Simple Spaghetti" recipe is correctly classified under MeatRecipe, due to the presence of an ingredient that is classified as a meat-based product in FOODON. The right hand panel shows OWL logical axioms for the recipe, including its ingredients, and the steps involved.
}
\label{figS3}
\end{figure*}

\begin{table*}[ht]
\caption{Resources used for grounding during evaluation of SPIRES with relations in the BC5CDR test corpus. These resources were used for initial annotation and are subsequently normalized to MeSH. Annotations from the Gilda text entity normalization tool are retrieved through its API (\url{http://grounding.indra.bio/apidocs}) using the Ontology Access Kit.}
\begin{tabular*}{\textwidth}
{@{\extracolsep{\fill}}llll@{\extracolsep{\fill}}}
\toprule
Entity type               & Resource                                     & Prefix   & Source                 \\
\midrule
Chemical & Medical Subject Headings 2022                & MESH     &  \citep{Lipscomb2000-zv}        \\
                          & Chemical Entities of Biological Interest     & CHEBI    &  \citep{Hastings2016-fl} \\
                          & National Cancer Institute Thesaurus          & NCIT     &  \citep{Sioutos2007-kg}  \\
                          & Mapping of Drug Names and MeSH 2022          & MDM      &  \citep{Lipscomb2000-zv}        \\
                          & DrugBank                                     & DRUGBANK &  \citep{Wishart2018-xr}  \\
                          & Gilda                                     & N/A      &  \citep{Gyori2022-xx}    \\
\midrule
Disease  & Medical Subject Headings 2022                & MESH     &  \citep{Lipscomb2000-zv}        \\
                          & Mondo Disease Ontology                       & MONDO    &  \citep{Mungall2017-ff}  \\
                          & Human Phenotype Ontology                     & HP       &  \citep{Kohler2021-ss}   \\
                          & National Cancer Institute Thesaurus          & NCIT     &  \citep{Sioutos2007-kg}  \\
                          & Human Disease Ontology                       & DOID     &  \citep{Schriml2019-hz}  \\
                          & Medical Dictionary for Regulatory Activities & MEDDRA   &  \citep{Brown1999-zq} \\
\bottomrule
\end{tabular*}
\label{tableS1}
\end{table*}

\begin{table*}[!hbt]
\caption{MeSH identifiers used to define value sets during evaluation of SPIRES with relations in the BC5CDR test corpus. All identifiers in this table were treated as root nodes of a hierarchy, i.e., the value sets include all child MeSH terms.}
\begin{tabular*}{\textwidth}
{@{\extracolsep{\fill}}lll@{\extracolsep{\fill}}}
\toprule
Entity type               & MeSH identifier                              & MeSH term \\
\midrule
Chemical & D602    & Amino Acids, Peptides, and Proteins                             \\
                           & D1685   & Biological Factors                                              \\
                           & D2241   & Carbohydrates                                                   \\
                           & D4364   & Pharmaceutical Preparations                                     \\
                           & D6571   & Heterocyclic Compounds                                          \\
                           & D7287   & Inorganic Chemicals                                             \\
                           & D8055   & Lipids                                                          \\
                           & D9706   & Nucleic Acids, Nucleotides, and Nucleosides                     \\
                           & D9930   & Organic Chemicals                                               \\
                           & D11083  & Polycyclic Compounds                                            \\
                           & D13812  & Therapeutics                                                    \\
                           & D19602  & Food and Beverages                                              \\
                           & D45424  & Complex Mixtures                                                \\
                           & D45762  & Enzymes and Coenzymes                                           \\
                           & D46911  & Macromolecular Substances                                       \\
\midrule
Disease  & D001423 & Bacterial Infections and Mycoses                                \\
                           & D001523 & Mental Disorders                                                \\
                           & D002318 & Cardiovascular Diseases                                         \\
                           & D002943 & Circulatory and Respiratory Physiological Phenomena             \\
                           & D004066 & Digestive System Diseases                                       \\
                           & D004700 & Endocrine System Diseases                                       \\
                           & D005128 & Eye Diseases                                                    \\
                           & D005261 & Female Urogenital Diseases and Pregnancy Complications          \\
                           & D006425 & Hemic and Lymphatic Diseases                                    \\
                           & D007154 & Immune System Diseases                                          \\
                           & D007280 & Disorders of Environmental Origin                               \\
                           & D009057 & Stomatognathic Diseases                                         \\
                           & D009140 & Musculoskeletal Diseases                                        \\
                           & D009358 & Congenital, Hereditary, and Neonatal Diseases and Abnormalities \\
                           & D009369 & Neoplasms                                                       \\
                           & D009422 & Nervous System Diseases                                         \\
                           & D009750 & Nutritional and Metabolic Diseases                              \\
                           & D009784 & Occupational Diseases                                           \\
                           & D010038 & Otorhinolaryngologic Diseases                                   \\
                           & D010272 & Parasitic Diseases                                              \\
                           & D012140 & Respiratory Tract Diseases                                      \\
                           & D013568 & Pathological Conditions, Signs and Symptoms                     \\
                           & D014777 & Virus Diseases                                                  \\
                           & D014947 & Wounds and Injuries                                             \\
                           & D017437 & Skin and Connective Tissue Diseases                             \\
                           & D052801 & Male Urogenital Diseases                                        \\
                           & D064419 & Chemically-Induced Disorders \\                                 
\bottomrule
\end{tabular*}
\label{tableS2}
\end{table*}

\begin{table*}[ht]
\caption{Results for named entity recognition evaluation of SPIRES on chemical and disease entities in the BC5CDR corpus. The chunking strategy was not used in this evaluation. Grounding was performed against MeSH only - further accuracy may be afforded by use of alternate ontology annotators such as CHEBI or MONDO for chemical and disease, respectively.}
\begin{tabular*}{\textwidth}
{@{\extracolsep{\fill}}lllll@{\extracolsep{\fill}}}
\toprule
Entity type               & Model                                     & F-score   & Precision & Recall                 \\
\midrule
Chemical & GPT-3.5-turbo                & 69.70     &  0.89  & 0.57        \\
          & GPT-4                & 73.69     &  0.85  & 0.65        \\

\midrule
Disease  & GPT-3.5-turbo                & 61.70     &  0.87  & 0.48        \\
          & GPT-4                & 69.70     &  0.88  & 0.56        \\

\bottomrule
\end{tabular*}
\label{tableS3}
\end{table*}

\end{document}